%% file: paper.tex
\documentclass[]{onethree}

\usepackage[toc,page,header]{appendix}
\usepackage{minitoc}
\usepackage{amsmath,amssymb}
\usepackage{bm}
\usepackage{wrapfig}
\usepackage{enumitem}
\usepackage[most]{tcolorbox}
\usepackage{longtable}
\usepackage{array}
\usepackage{placeins}
\newtcolorbox{promptbox}[1]{
  enhanced,
  breakable,
  colback=black!2,
  colframe=black!25,
  colbacktitle=black!7,
  coltitle=black,
  fonttitle=\small\bfseries,
  fontupper=\small\rmfamily,
  title={#1},
  boxrule=0.4pt,
  arc=1mm,
  left=8pt,right=8pt,top=6pt,bottom=6pt,
  before skip=10pt,after skip=10pt
}

\title{Learning from Physical Experience:\\
Context-Space Adaptation for Robotic Manipulation}

\author[1,2,3]{Yihan Zeng}
\author[1]{Yisu Li}
\author[1]{Kaixin Huang}
\author[2]{Marc WuDunn}
\author[2]{Amelia Kovacs}
\author[2]{Mikael Harseno}
\author[2]{Kimmy Lin}
\author[2]{Porter Hunley}
\author[2]{Fred Shentu}
\author[2]{Philipp Wu}
\author[1]{Zhongyu Li}
\author[1]{Zike Yan}

\renewcommand{\affiliationlist}{%
  \affiliationformat[1]{The Chinese University of Hong Kong},
  \affiliationformat[2]{xdof.ai}\\
  \affiliationformat[3]{University of Electronic Science and Technology of China}%
}

\abstract{
Recent multimodal foundation models exhibit strong perception, reasoning, and in-context learning capabilities, but translating these capabilities into physical action remains challenging.  Through real-world experiments across diverse manipulation tasks, we find that ChatGPT-6-Astra, a state-of-the-art general-purpose reasoner, can serve as a physically actionable, context-responsive manipulation policy, demonstrating effective fine manipulation and spatial reasoning. Building on these findings, we introduce LEXP, a harness for learning from physical experience through context-space adaptation. LEXP combines task-level planning with an experience module that interprets execution evidence, extracts and retains lessons with applicability conditions, and reuses relevant lessons to generate contextual feedback for subsequent actions. Model weights and the action interface remain fixed throughout. Across 12 real-world manipulation tasks, LEXP achieves $23/36$ successful trials (63.9\%), compared with $19/36$ (52.8\%) with memory alone and $2/36$ (5.6\%) for vanilla Astra. On three tasks, bread transfer, sort trash and toiletries, and retrieve drawer, removing planning reduces success from $7/9$ to $4/9$, while removing retained experience reduces it to $3/9$. Execution traces illustrate feedback-guided corrections and cross-episode reuse of an inspection strategy. Execution analyses illustrate how task structure and experience-derived feedback guide task progression, action correction, and experience reuse. These results support context-space adaptation as a practical mechanism for turning physical interaction into reusable guidance for robotic manipulation.
\vspace{-0.1cm}
}

\checkdata[Resources]{
  \href{https://hatty-z.github.io/LearnFromExperience/}{\raisebox{-1ex}{\includegraphics[height=4.2ex]{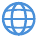}}~Project Page}\hspace{2em}
}

\fancypagestyle{firststyle}{
    \fancyhf{}
    \fancyfoot[C]{\thepage}
}

\begin{document}
\maketitle

\input{figures/teaser}

\input{sections/introduction}
\input{sections/problemstatement}
\input{sections/modeldesign}
\input{sections/experiments}
\input{sections/related_works}
\input{sections/conclusion}

\clearpage

\bibliographystyle{plainnat}
\bibliography{main}

\clearpage

\beginappendix
\input{sections/appendix}

\end{document}

%% file: figures/teaser.tex
\begin{figure}[!t]
    \includegraphics[width=\textwidth]{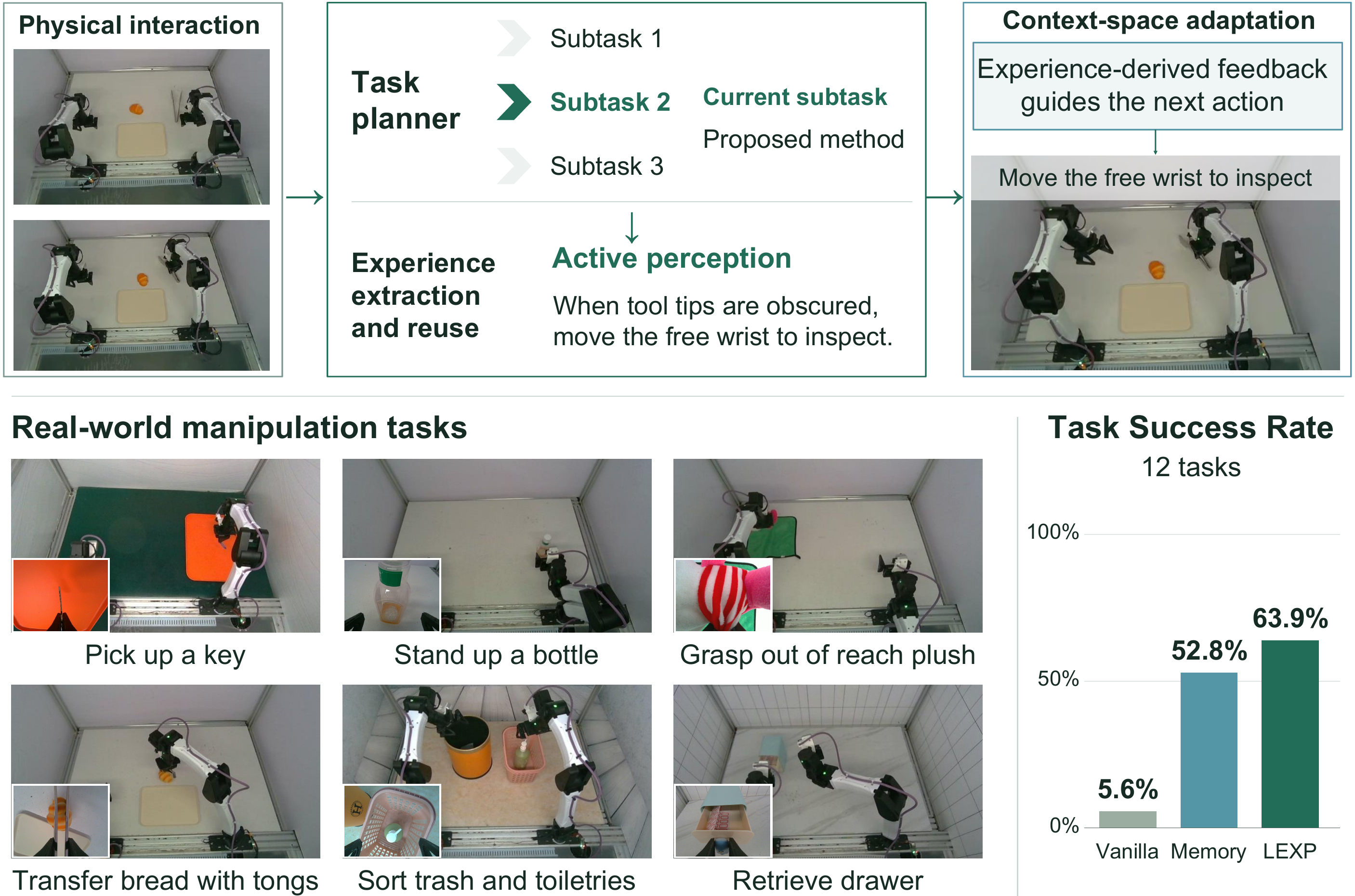}
    \caption{
    \textbf{Learning from physical experience through context-space adaptation.}
    LEXP combines task-level planning with experience extraction and
    reuse to guide subsequent actions while keeping model weights and
    the action interface fixed. Across 12 real-world manipulation tasks,
    LEXP achieves 63.9\% success, compared with 52.8\% with memory alone
    and 5.6\% for vanilla Astra.
    }
    \label{fig:teaser}
\end{figure}

%% file: sections/introduction.tex
\section{Introduction}

A long-standing goal of robotic manipulation is to perform diverse tasks reliably in unfamiliar and changing physical environments. This requires robots to adapt when familiar strategies fail and
reuse what they learn from physical interaction. The challenge is to turn limited physical experience into reliable, transferable improvements in behavior. Existing approaches pursue this goal through different adaptation mechanisms. Vision--language--action (VLA) models learn from robot demonstrations through supervised fine-tuning \citep{openvla,pi05}. World action models  \citep{dreamzero}aim to reduce task-specific adaptation by jointly predicting video and actions, enabling zero-shot execution of unseen tasks and environments on pretrained embodiments, while related video-model policies \citep{cosmospolicy} instead use demonstration-based fine-tuning and rollout-informed planning . Real-world reinforcement learning improves policy parameters through repeated physical interaction \citep{pi06}, but requires reward specification and infrastructure
for physical trials and resets \citep{pi06,serl,hilserl}. These requirements make demonstrations, reward engineering and physical trial infrastructure central to the cost of adaptation. General-purpose multimodal reasoners offer a complementary interface: observations, instructions and execution feedback can condition executable tool calls without updating model weights. Their relevance to robotics depends on whether this contextual flexibility supports effective action revision and reusable improvements under a limited physical interaction budget.

We investigate this possibility through real-world experiments with GPT-6 Astra \citep{openai2026astra}, across 12 robotic manipulation tasks covering fine manipulation, spatial reasoning, long-horizon execution, tool use, bimanual coordination, and force control. Across these tasks, we vary access to interaction history, and examine recovery demonstrations and online language guidance. Two complementary findings emerge. First, Astra shows promising fine-manipulation and spatial reasoning capabilities when given access to interaction history: interaction history inform subsequent corrections and help the robot reason about object relations and pose. Second, its behavior is steerable through context. Visual demonstrations and textual corrections support revisions to grasp checking, object repositioning, motion direction, arm selection, and action order. These benefits vary across tasks, but together highlight both Astra's executable competence and its capacity for context steering---two prerequisites for improving manipulation through experience-derived context rather than additional parameter training.

Building on these observations,we introduce LEXP, a harness for learning from physical experience through context-space adaptation (Figure~\ref{fig:teaser}). A context compiler combines task instructions, current observations, interaction history, and agent-generated experience feedback to condition executable actions. A planner decomposes the task into subtasks, tracks progress, and proposes a method for the current subtask. The experience module evaluates this method against recent actions, measured execution outcomes, and current observations, drawing on retained lessons to generate feedback for the actor's next decision. It also extracts and updates structured lessons from physical interaction, recording their applicability, supporting evidence, and limitations for subsequent reuse. This separates interaction history, which preserves past exchanges, from reusable experience, which retains findings beyond the history window and across attempts. The models remain frozen and the action interface stays fixed; adaptation occurs through changes to the context guiding execution. The design builds on reflective context optimization \citep{gepa,tfgrpo} and prior physical adaptation systems \citep{aspire,enpire,capx,harnessvla}.

Across 12 real-world manipulation tasks, LEXP achieves $23/36$ successful trials, compared with $19/36$ with memory alone and $2/36$ for vanilla Astra. Ablations and execution analyses further examine how task-level planning and retained experience
support task progression and subsequent action decisions.

Our contributions are threefold:
\begin{itemize}
    \item A real-world evaluation suite spanning 12 manipulation tasks to investigate executable capabilities, adaptation through interaction history, and context steering in a general-purpose multimodal reasoner. Across three fine-manipulation tasks, memory increases Astra's success rate from $0/9$ to $8/9$.
    \item LEXP, a harness that combines task-level planning with the extraction and reuse of physical experience to guide robotic manipulation through context-space adaptation. Across 12 tasks, LEXP achieves $23/36$ successful trials, compared with $19/36$ with memory alone and $2/36$ for vanilla Astra.
    \item Component ablations and execution analyses examining the roles of planning and retained experience. Across three tasks, bread transfer, sort trash and toiletries, and retrieve drawer, complete LEXP achieves $7/9$ successful trials, compared with $4/9$ without planning and $3/9$ without retained experience.
\end{itemize}


%% file: sections/problemstatement.tex
\section{Problem Statement}

\subsection{Objective: improve policy behavior through reusable experience}

Our objective is to improve robotic manipulation by converting physical experience into reusable guidance for subsequent behavior. Supervised fine-tuning encodes demonstrated actions in model parameters, while reinforcement learning adjusts a policy using rewards from interaction. We pursue the same practical goal to improve performance but move the variables being optimized from model parameters to the context that conditions the policy. With model weights and the robot interface fixed, contextual
guidance can change how the actor advances the current subtask,
approaches an object, verifies a grasp, and responds to failure.

\textbf{Context-space Adaptation.}
Let $\pi_\theta$ denote a pretrained actor with fixed parameters
$\theta$, and let $\mathcal{I}$ denote its fixed robot-control
interface. Given a task instruction $l$ and a current observation
$o_t$, the actor selects an action conditioned on a compiled
context:
\begin{equation}
    a_t \sim \pi_\theta(\cdot \mid c_t),
    \qquad
    c_t = \mathcal{C}(l, o_t, H_t, f_t),
    \label{eq:context_policy}
\end{equation}
where $H_t$ is memory consisting of interaction history,
$f_t$ is contextual guidance, and $\mathcal{C}$ is the
context compiler. Executing $a_t$ through $\mathcal{I}$ produces an execution
record $r_t$ and a subsequent observation $o_{t+1}$.
The resulting evidence
$e_t = (o_t, a_t, r_t, o_{t+1})$
records the proposed action, its execution result, and the
observed physical consequences.

We implement this adaptation through structured, modular
context updates \citep{ace}.
After each interaction, new evidence is appended to memory:
$H_{t+1} = \operatorname{Append}(H_t, e_t)$.
Let $M_t$ denote retained experience and $p_{t+1}$ denote the
planner's current subtask and proposed method.
The experience module $\mathcal{E}$ uses this task-level
reference together with memory, the current observation,
and retained experience to update lessons and generate feedback:
\begin{equation}
\begin{aligned}
    (M_{t+1}, f_{t+1})
        &= \mathcal{E}(o_{t+1}, H_{t+1}, p_{t+1}, M_t), \\
    c_{t+1}
        &= \mathcal{C}(l, o_{t+1}, H_{t+1}, f_{t+1}).
\end{aligned}
\label{eq:context_adaptation}
\end{equation}
Planning and retained experience thus inform the actor through
generated feedback.
Physical execution supplies new evidence, which updates memory,
retained experience, and feedback; the context compiler then
assembles the next actor input, closing the adaptation loop.
Retained experience remains available across attempts, allowing
lessons to be reused beyond their source interactions.
Module-specific prompts are provided in
Appendix~\ref{app:prompts}.


\subsection{Preliminary capability and context-use evidence of GPT-6 Astra}
\label{sec:preliminary}

We investigate the prerequisites for context-space adaptation through real-world experiments with GPT-6 Astra \citep{openai2026astra} across 12 robotic manipulation tasks covering fine manipulation, spatial reasoning, long-horizon execution, tool use, bimanual coordination, and force control. We compare execution with and without interaction history and separately examine the effects of recovery demonstrations and online language feedback. These experiments examine two empirical prerequisites for
context-space adaptation: whether memory enables the actor
to use physical interaction evidence in subsequent decisions,
and whether contextual guidance can steer its behavior.
\input{figures/memory_results}

\textbf{Memory benefits fine manipulation through outcome-aware corrections.}
Figure~\ref{fig:preliminary_capability}a compares Astra's small-object pickup performance with and without memory. Overall success rises from $0/9$ (0\%) without memory to $8/9$ (88.9\%) with memory. The task-level contrasts are $0/3$ versus $3/3$ for cable ties, $0/3$ versus $3/3$ for keys, and $0/3$ versus $2/3$ for the hex key. Here, memory consists of interaction history carried into the actor's context, including previous observations, actions and their intention notes, and execution outcomes.

These results show that Astra can execute precision-demanding grasps and that memory guidance can help it deploy this capability successfully which allows the actor to interpret the current scene in relation to earlier attempts, assess what an action changed, and use that evidence to revise the next command. Poor performance without memory  therefore does not imply that the required manipulation is beyond its execution capability. Instead, the contrasts motivate studying how prior observations and action outcomes guide subsequent choices. This benefit is actor-dependent: across the same three tasks, Opus 5.5 improves from $0/9$ to $4/9$, whereas Sol 5.6 changes from $1/9$ to $0/9$ (Figure~\ref{fig:preliminary_capability}b--c). The observations support experience-guided adaptation, while leaving open which history contents produce the gains.

\textbf{Memory supports spatial reasoning about object relations and pose.}
Across two spatial manipulation tasks, Astra's observed success increases from $0/3$ without memory to $3/3$ with memory for grasping an out-of-reach plush, and from $0/3$ to $2/3$ for standing a fallen bottle upright. The plush task involves reasoning about reachability and object relations, while bottle reorientation requires tracking changes in orientation and support. By retaining previous observations, actions, and execution outcomes, memory makes the object's response to manipulation available to subsequent decisions, helping the actor interpret spatial configurations through interaction rather than appearance alone.

\input{figures/context_steering}
\textbf{Context steering changes strategies and follows spatially specific instructions.}
Visual demonstrations and online language feedback steer behavior at both the strategy and motion levels. In towel folding, a clean demonstration leads Astra to grasp the near corners, retry the fold, and eventually give up, whereas a recovery demonstration leads it to verify edge capture, maintain the grasp, bring the towel closer, and inspect the resulting fold (Figure~\ref{fig:steering}a). The difference concerns grasp verification and workspace management, showing that demonstration context can redirect the manipulation strategy.

Figure~\ref{fig:steering}b illustrates online instruction following during barbecue transfer. A request to move backward is followed by a 4.4\,cm displacement; feedback requesting more clearance produces a 12.8\,cm lift; and guidance to clear the left arm produces a 12.0\,cm left-arm motion while the right remains fixed. ``Up, then right'' guidance leads to a 2.7\,cm rise followed by a 4.2\,cm lateral motion. These examples demonstrate responsiveness to motion direction, arm selection, and action order. Visual demonstration context and textual feedback thus provide concrete channels for revising both high-level strategies and individual actions, motivating our harness design to deliver guidance derived from physical experience through context.

\textbf{From context use to experience-derived adaptation.}
These observations motivate a harness that turns physical interaction into reusable contextual guidance. Memory makes past interactions available, while demonstrations and language feedback show that context can steer subsequent behavior. The next question is whether useful guidance can be extracted from the actor's own successful and failed interactions and reused in new executions.

%% file: figures/memory_results.tex
\begin{figure}[!t]
\centering\includegraphics[width=\linewidth]{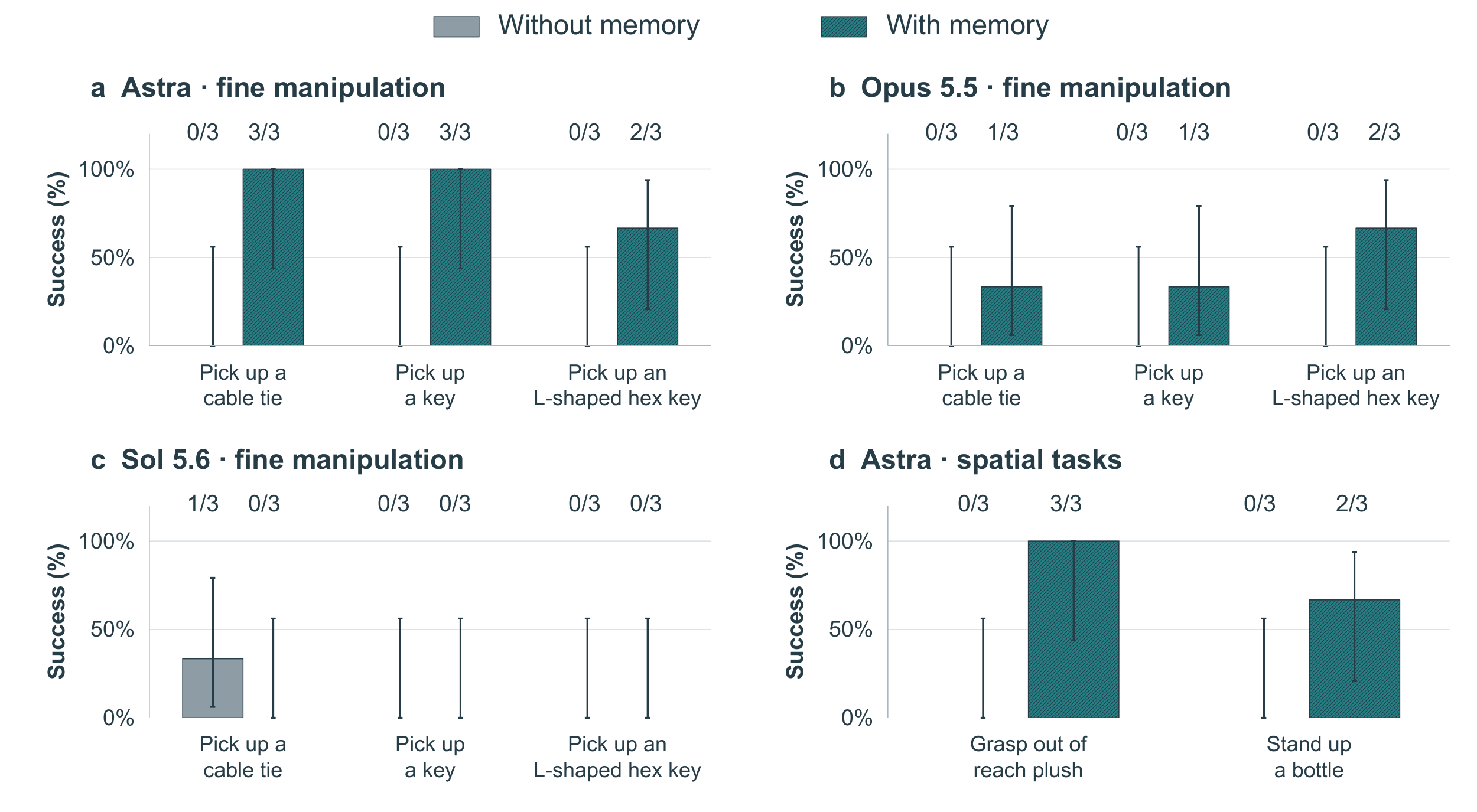}
\caption{Memory supports fine manipulation and spatial reasoning. Success rates with and without memory for (a) GPT-6 Astra, (b) Opus 5.5, and (c) Sol 5.6 on three fine-manipulation tasks, and (d) GPT-6 Astra on two tasks requiring spatial reasoning. Memory consists of prior observations, actions and intention notes, and execution outcomes retained in the actor's context. Bars show observed success rates, with successful trials over total trials indicated above each bar; error bars denote 95\% Wilson confidence intervals. Memory improves Astra's performance across the evaluated tasks.}
\label{fig:preliminary_capability}
\end{figure}

%% file: figures/context_steering.tex
\begin{figure}[!ht]
\centering
\includegraphics[width=\linewidth]{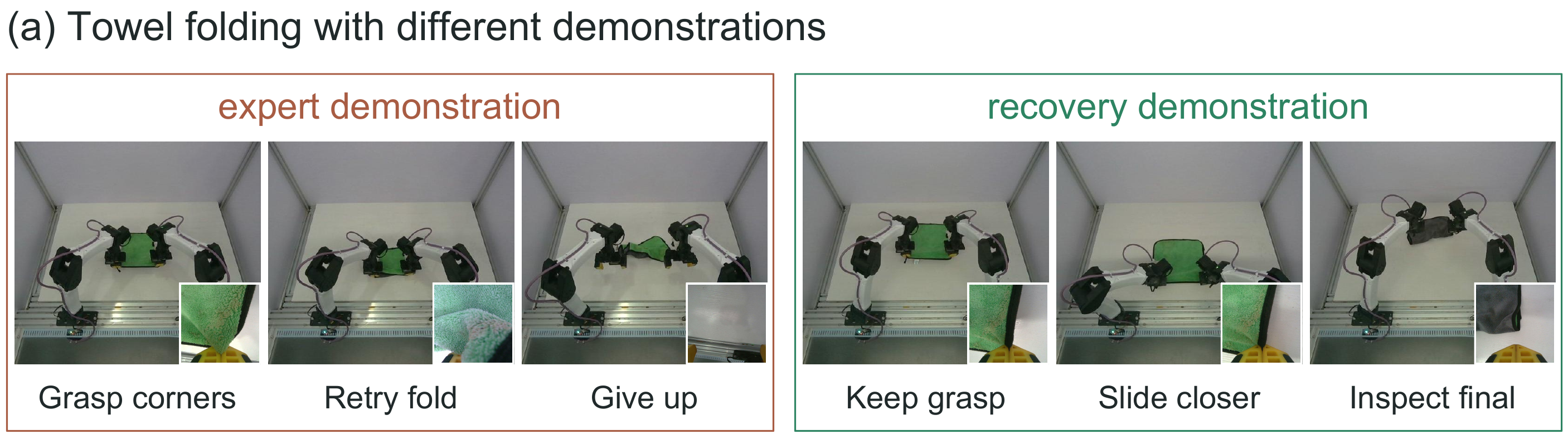}\\[3pt]
\includegraphics[width=\linewidth]{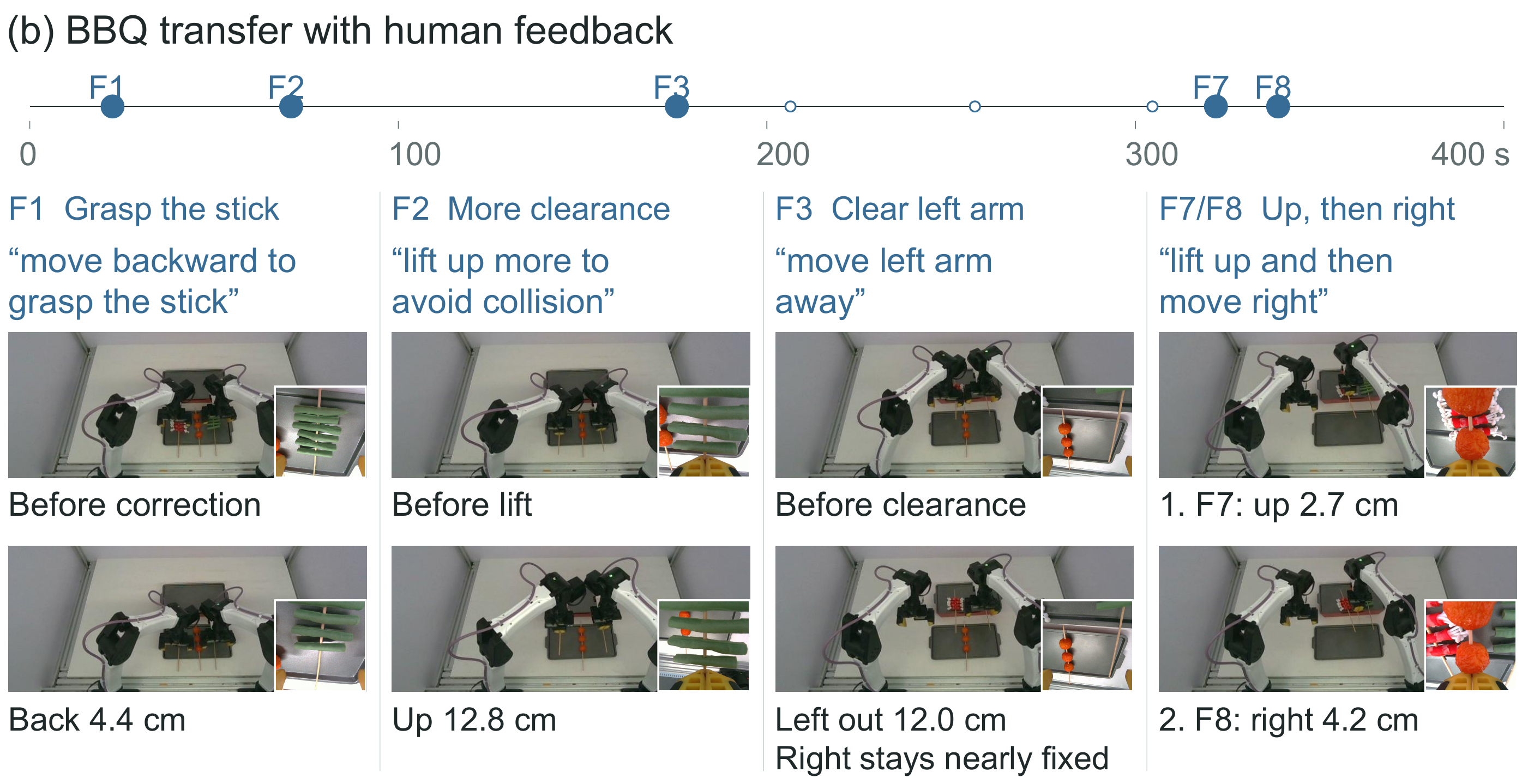}
\caption{Context steering changes strategies and follows spatially specific instructions. (a) Expert and recovery demonstrations lead to different grasp-checking and repositioning strategies during towel folding. (b) During barbecue transfer, online language feedback guides changes in grasp approach, lifting clearance, arm selection, and motion order. Selected observations and measured displacements illustrate the robot's responses, including moving backward to grasp, lifting to avoid collision, moving the left arm away, and lifting before moving right. The timeline marks feedback delivery; F labels identify individual feedback messages.}
\label{fig:steering}
\end{figure}

%% file: sections/modeldesign.tex
\section{Method}

Building on these observations, we introduce LEXP, a harness for learning from physical experience through context-space adaptation. As shown in Figure~\ref{fig:method_overview} , LEXP combines executable action decisions, task-level planning, and experience-derived feedback to turn physical interactions into reusable guidance, while keeping model weights and the action interface fixed. We first present the framework and its information flow in Section~\ref{sec:method_overview}. We then describe how the actor's context is constructed and its actions are executed in Section~\ref{sec:method_actor}, followed by task decomposition and progress tracking in Section~\ref{sec:method_planner}. Finally, Section~\ref{sec:method_experience} details how physical experience is extracted, retained, and reused to generate feedback for subsequent actions.



\subsection{Overview}
\label{sec:method_overview}
Figure~\ref{fig:method_overview} illustrates LEXP's interaction loop. At each decision, a context compiler assembles the system prompt, task instruction, current visual observations and robot states, memory, and available guidance for the actor. The actor issues a tool call specifying an action and an accompanying intention note. The execution layer checks the proposed action before execution and records its outcome. Subsequent observations capture the resulting robot state and scene, completing an interaction record for later decisions.

Alongside this loop, the planner decomposes the task into subtasks, assesses progress from the observed scene, and proposes a method for the current subtask. These outputs inform the experience module. The experience module checks the proposed method against recent interactions and current observations, applies relevant retained lessons, and generates feedback for the actor's next action. It also extracts and revises lessons as new evidence becomes available, preserving their applicability conditions, supporting evidence, and limitations. Physical interaction thus informs future behavior through both raw history and reusable experience, with the latter translated into contextual feedback for the current decision.

\input{figures/method_overview}

\subsection{Actor Context and Action Execution}
\label{sec:method_actor}
The context compiler assembles the actor's input from the system prompt, task instruction, current camera observations and robot states, memory and available guidance. Memory contains the proposed action and its intention note, execution outcome, and observations before and after the action, allowing the actor to relate the current scene to previous attempts. The planner's proposed method and retained experience are interpreted by the experience module and reach the actor through its feedback.

At each decision, the actor produces one tool call containing a motion command and an intention note, or signals completion or termination. The execution layer validates proposed motions against the latest robot state and execution constraints, then executes accepted commands. The command, execution outcome, and subsequent observations are recorded as a completed interaction. These records provide evidence of the action's physical effects for both the actor's history and the experience module.

\subsection{Task-Level Planning}
\label{sec:method_planner}

Planning provides task-level structure for reasoning and execution. Individual action decisions must serve the overall task, while retained lessons must be interpreted in relation to what the robot is currently trying to accomplish. The planner supplies this structure by decomposing the task into an ordered list of subtasks. As execution proceeds, it assesses progress against the observed scene, identifies the current subtask, and proposes a method for carrying it out. Together, the current subtask and proposed method give the experience module a reference for assessing recent execution and applying relevant lessons to the current situation. Through the resulting feedback, planning guides task-level reasoning and the application of experience.

\subsection{From Physical Experience to Context Space Adaptation}
\label{sec:method_experience}
The experience module turns physical interaction into contextual guidance by combining recent execution evidence with retained lessons. It receives the current subtask and proposed method from the planner, recent interaction records, and current observations. It first checks whether the proposed method's condition is satisfied and compares observed execution with the method. This assessment grounds feedback in what the robot commanded and what the observations show, rather than assuming that an accepted command achieved its intended physical effect.

\paragraph{\textbf{Extracting and retaining experience}.}
Alongside feedback generation, the module records useful findings from completed interactions. These may include observed object relations, attempted methods and their results, partial successes, unsuccessful attempts, and execution constraints. Each lesson is retained with its applicability conditions, its potential use in later decisions, supporting interaction records, and limitations or unresolved explanations. New evidence can revise an existing lesson, allowing retained experience to change as the robot encounters additional physical evidence. Unlike raw history, these entries preserve selected findings after the original interactions leave the actor's history window.

\paragraph{\textbf{Reusing experience through contextual feedback.}}
For subsequent decisions, the module considers lessons retained from the current attempt or carried over from earlier attempts. It evaluates their applicability against the current observations, using the planner's subtask and proposed method as a task-level reference. Relevant lessons inform a concise feedback message supplied to the actor before its next action; when retained experience conflicts with current observations, the observations take precedence. Experience reuse therefore involves interpreting prior findings in the current physical situation and translating them into actionable contextual guidance.

%% file: figures/method_overview.tex
\begin{figure}[t]
    \centering
    \includegraphics[width=0.95\linewidth]{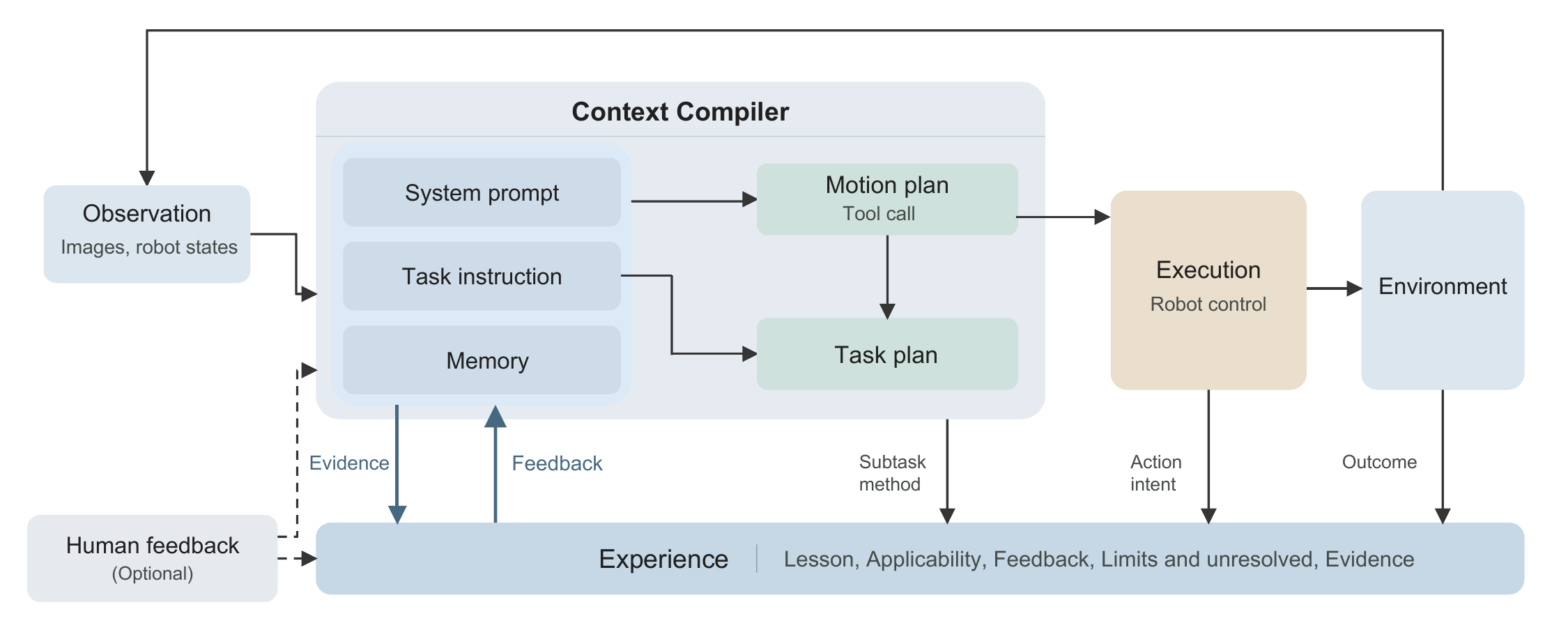}
    \caption{\textbf{Method overview.} The context compiler combines task instructions, current observations and robot states, memory, and experience-derived feedback to condition the actor's next action. Physical execution produces outcomes and subsequent observations that become evidence for later decisions. The planner tracks task progress and proposes a method for the current subtask. The experience module interprets this task-level structure alongside physical evidence and retained lessons, generating contextual feedback while extracting and revising reusable experience. Model weights and the action interface remain fixed throughout.}
    \label{fig:method_overview}
\end{figure}

%% file: sections/experiments.tex
\input{figures/task_overview}
\section{Experiments}
\label{sec:evaluation}

We evaluate LEXP through real-world manipulation experiments organized around four questions: \textbf{Q1}, Does LEXP improve performance on real world manipulation tasks? \textbf{Q2}, How does task-level planning guide the actor's reasoning and task progression? \textbf{Q3}, Can lessons extracted from physical interaction guide later decisions beyond the interactions from which they were learned? \textbf{Q4}, How does experience-derived feedback shape the actor's action decisions?

\subsection{Experimental Setup}
\label{sec:experimental_setup}

\paragraph{Platform and tasks.}
We conduct experiments on a bimanual YAM robot equipped with two grippers and three view realsense cameras. The actor receives camera observations and robot states and controls the robot through a tool-call interface. Our evaluation covers 12 manipulation tasks spanning fine manipulation, spatial reasoning, long-horizon execution, tool use, bimanual coordination, and force control. Figure~\ref{fig:task_overview} illustrates the 12 tasks and representative stages of their execution. Task details are provided in Appendix \ref{app:task_details}. The preliminary experiments in Section~\ref{sec:preliminary} characterize memory and context steering; here, we evaluate LEXP's manipulation performance and examine the roles of task-level planning and experience reuse.

\paragraph{Comparison conditions.}
We compare the same actor without memory, with memory, and with LEXP. The memory condition retains previous observations, actions and intention notes, and execution outcomes. LEXP additionally uses task-level planning and experience-derived feedback as described in Sections~\ref{sec:method_planner}
and~\ref{sec:method_experience}. Model weights and the action interface remain fixed across conditions.

\paragraph{Protocol and metrics.}
For each task, we conduct three trials. Trials across comparison conditions use similar distributions of initial configurations. Retained experience is initialized according to the evaluated condition, with any experience carried over from earlier attempts explicitly identified. Each trial is limited to 100 actor decisions and ends when the actor signals completion, or when the budget is exhausted. Task success is assessed by human annotations using predefined physical completion criteria, independently of the actor's completion declaration. We report success as the number of successful trials out of three and subtask progress as the total number of completed annotated subtasks divided by the total number of evaluated subtasks across trials, using annotations independent of the planner’s generated checklist. Task-specific initialization procedures and success criteria are provided in Appendix~\ref{app:task_details}.

\input{figures/overall_performance}

\subsection{Q1: Does LEXP improve performance on real world manipulation tasks?}

To answer Q1, we compare vanilla Astra, Astra with memory, and Astra equipped with LEXP across 12 real-world manipulation tasks. As shown in Figure~\ref{fig:overall_performance}a, LEXP achieves
$23/36$ successful trials (63.9\%), compared with $19/36$ (52.8\%) with memory and $2/36$ (5.6\%) without memory. Retaining interaction history substantially improves observed performance, while LEXP provides a further aggregate gain of 11.1 percentage points over memory alone.
Complete task-level results, including the other evaluated models, are reported in Appendix~\ref{app:complete_results}, Table~\ref{tab:full_manipulation_results}.

Figure~\ref{fig:overall_performance}c shows that the additional benefit over memory is concentrated in long-horizon execution, where success increases from $3/12$ to $7/12$. Tool-use and force-control tasks also show higher aggregate success, while fine-manipulation and spatial-reasoning performance remains unchanged in aggregate. Bimanual coordination remains challenging, with no successful sweeping or towel-folding trials under any condition. The following analyses examine how planning, feedback, and
retained experience influence execution.

\input{figures/ablation}

\subsection{Q2: How does task-level planning guide the actor's reasoning and task progression?}

Planning provides task-level structure for reasoning and execution. To examine its contribution, we remove the planner while retaining memory, the experience module, and experience-derived feedback. Across bread transfer, drawer retrieval, and item sorting, complete LEXP achieves
$7/9$ successful trials and $24/27$ completed subtasks, compared with $4/9$ and $16/27$ without the planner (Figure~\ref{fig:ablation}).

The actor can already reason about how to complete these tasks, but execution requires maintaining task objectives through repeated actions and recovery. Different tasks impose different demands, bread transfer follows a relatively direct acquisition then transfer then placement sequence, whereas sorting requires preserving item and container assignments and drawer search requires tracking inspected drawers and opening constraints. Item sorting illustrates the distinction between initial task reasoning and its sustained use. In the memory-only example, the actor initially assigns the bottle to the trash bin and later releases the paper into the pink basket. Without the planner, experience-derived feedback already provides task-level guidance: before the first action, it asks the actor to establish the sorting rule rather than infer destinations from proximity. The actor subsequently states the correct item and container assignments in its action
note. Later, feedback recognizes that the paper is inside the bin and redirects attention to the remaining bottle, demonstrating implicit progress tracking. Although the actor revisits the container choice during the prolonged execution and again selects the basket, bottle placement remains incomplete.

LEXP provides an explicit task-level reference through separate subtasks for placing paper in the orange bin and the bottle in the pink basket. The experience module evaluates execution against these goals. When the paper remains caught on an open fingertip, feedback directs release recovery over the bin and requires visible separation before counting placement as complete. Once paper placement is confirmed, execution proceeds to the bottle, which is subsequently released inside the basket. These traces show that experience feedback can provide task-level guidance on its own, while explicit planning structures this guidance around identified subtasks and their observed completion.

\input{figures/plan_experience}

\subsection{Q3: Can lessons extracted from physical interaction guide later decisions beyond the interactions from which they were learned?}

Physical experience becomes reusable guidance in context space. To examine the contribution of retained experience, we compare
complete LEXP with a variant that retains planning and immediate
feedback but does not retain or access stored lessons. Across the
three evaluated tasks, complete LEXP achieves $7/9$ successful
trials and $24/27$ completed subtasks, compared with $3/9$ and
$12/27$ without retained experience
(Figure~\ref{fig:ablation}). Observed task success is higher with
retained experience on all three tasks. This comparison supports
the utility of explicit lesson retention and reuse in the evaluated
settings.

Figure~\ref{fig:experience_reuse} illustrates the development and
reuse of an inspection strategy across three bread-transfer episodes.
In Episode~1, the actor recognizes that the bread lies between the
tong shafts and retracts and reorients
the tool to improve tip visibility in the top-camera view. The
selected inspection action begins approximately 333\,s after the
first actor call. Retained experience records when this strategy
applies and the need to verify tip alignment and bread co-motion.
In Episodes~2 and~3, the actor executes a free-wrist camera
repositioning action approximately 88\,s and 83\,s after the first
actor call, respectively. The retained lesson is subsequently
refined to include free-wrist inspection when tool-hand adjustments
do not resolve visibility. These traces illustrate inspection
behavior and experience reused across episodes.

\subsection{Q4: How does experience-derived feedback shape the actor's action decisions? }

To answer Q4, we compare memory-only execution with complete LEXP and examine how feedback informs subsequent actions. Across the three tasks in Figure~\ref{fig:ablation}, LEXP achieves $7/9$ successful trials and $24/27$ completed subtasks,
compared with $3/9$ and $13/27$ with memory alone. This comparison establishes the performance benefit of the complete harness; execution traces reveal how task structure and retained experience reach the actor through contextual guidance.

The sorting and bread-transfer examples illustrate two forms of feedback uptake.
In sorting (Figure~\ref{fig:planning_sort_comparison}), the experience module identifies that paper remains caught on a fingertip despite the jaws being open. Its feedback directs an orientation adjustment and requires visible separation before counting placement as complete. The actor subsequently adjusts the wrist, and later
observations confirm the paper inside the bin, separate from the fingers. The bread-transfer traces (Figure~\ref{fig:experience_reuse})
provide a complementary example of experience reuse decisions.
The recorded action intentions explicitly call for exposing the
distal tong tips: Episode~1 uses tool reorientation, while
Episodes~2 and~3 reposition the free wrist camera for an independent
view. These actions correspond to the inspection strategies
represented in retained experience.  Together, these traces show feedback guiding both physical recovery and information-gathering actions, translating task-level objectives and prior experience into changes in the actor's subsequent behavior.

%% file: figures/task_overview.tex
\begin{figure*}[p]
    \centering
    \includegraphics[width=\textwidth]{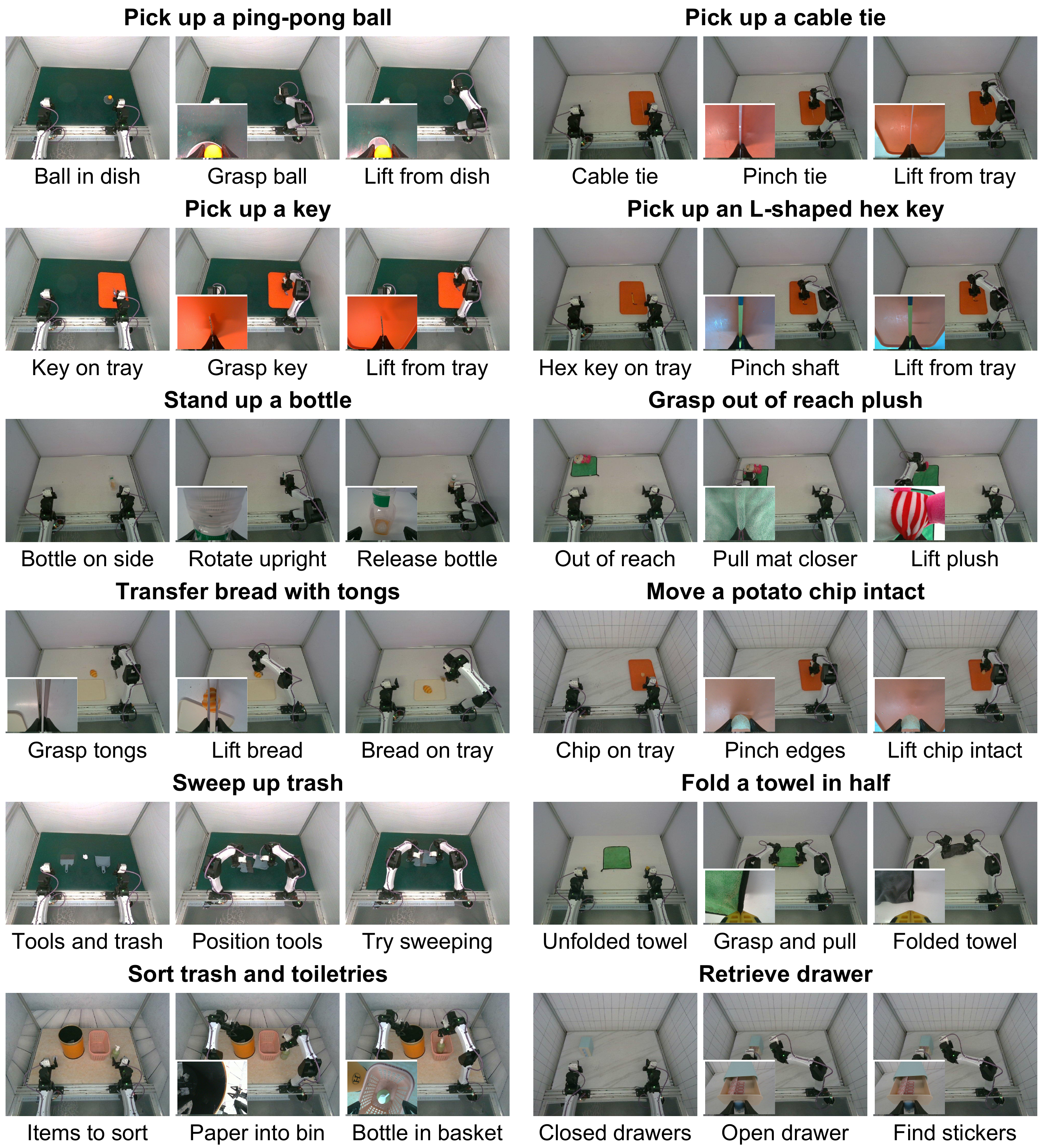}
    \caption{\textbf{Overview of the 12 real-world manipulation tasks.}
    Each task is illustrated by three selected keyframes showing its
    initial configuration and representative physical interactions.
    Main images show top-camera observations; insets provide wrist-camera
    views of grasping, tool use, and object placement.
    The sequences illustrate task content and execution stages;
    quantitative performance is reported separately.
    Task instructions and completion criteria are provided in
    Appendix~\ref{app:task_details}.}
    \label{fig:task_overview}
\end{figure*}

%% file: figures/overall_performance.tex
\begin{figure*}[t]
    \centering
    \includegraphics[width=\textwidth]{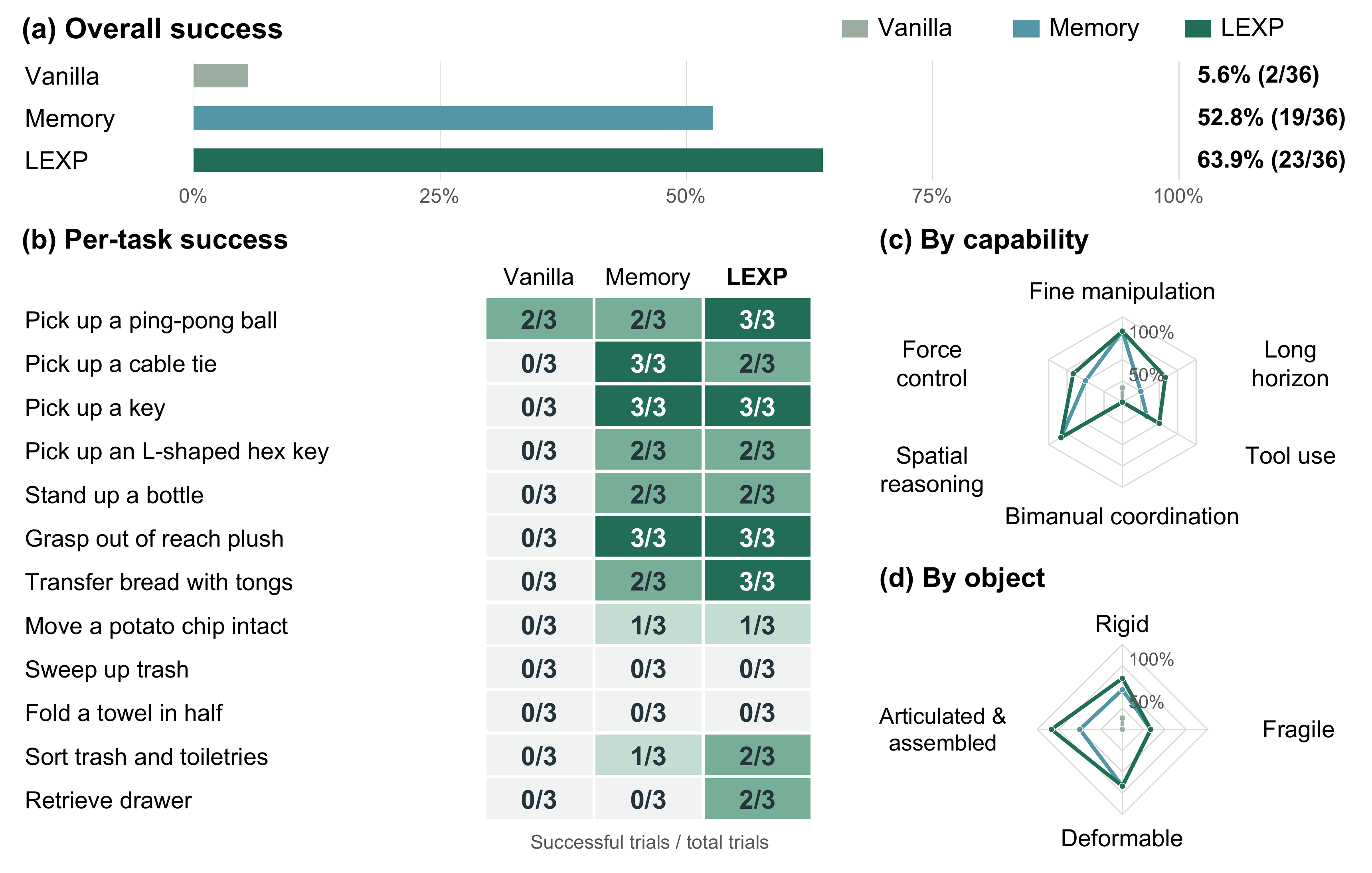}
    \caption{\textbf{Overall performance comparison.}
    We compare vanilla Astra, Astra with memory, and Astra equipped
    with LEXP.
   (a) Overall success across 12 tasks; labels indicate successful
trials over total trials.
(b) Per-task success; each cell reports successful trials out
of three, with darker shading indicating higher success.
(c) Mean task-success rates by capability.
(d) Mean task-success rates by primary object category.
Capability categories may overlap, while each task belongs to
one primary object category. Each task is evaluated in three
trials per condition. Complete task-level results are reported
in Table~\ref{tab:full_manipulation_results}.}
    \label{fig:overall_performance}
\end{figure*}

%% file: figures/ablation.tex
\begin{figure*}[t]
    \centering
    \includegraphics[width=\textwidth]{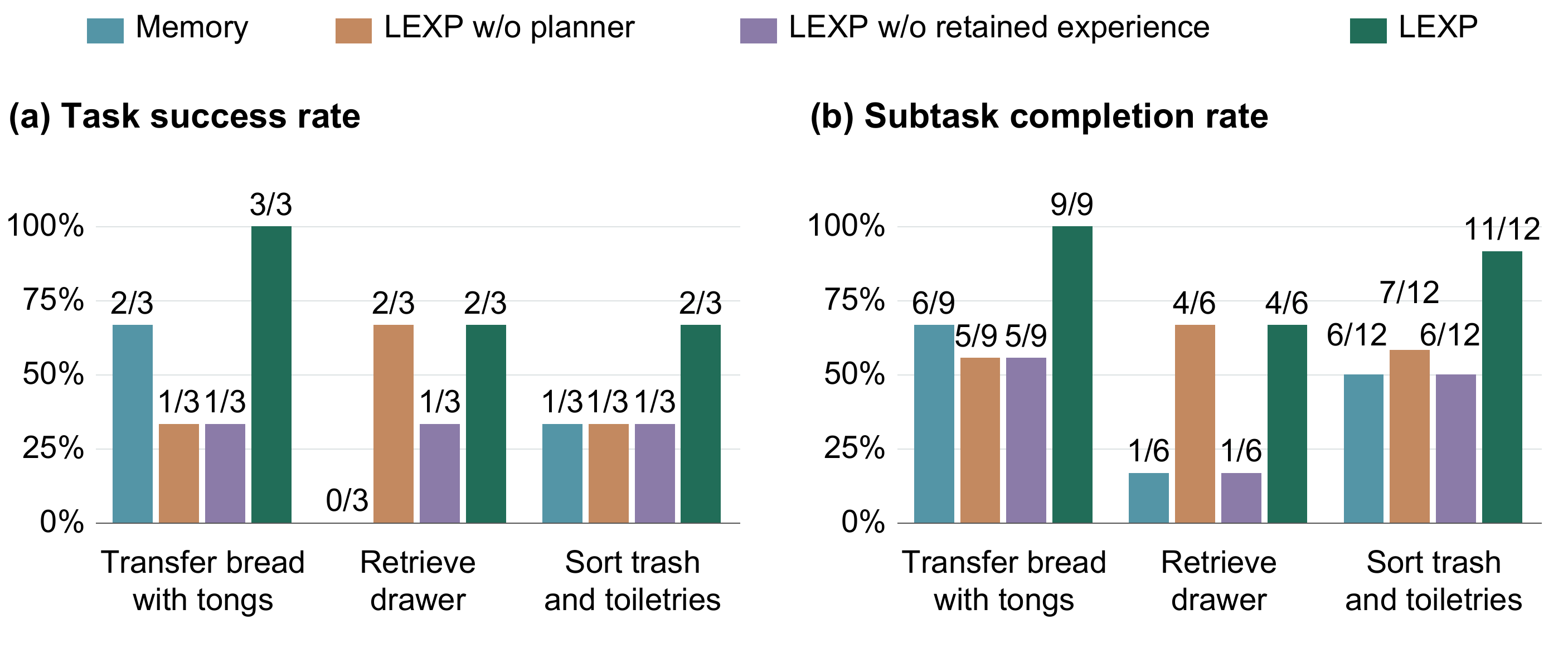}
    \caption{\textbf{Effects of planning and retained experience on
    manipulation performance.}
    We compare memory alone, LEXP without the planner, LEXP without
    retained experience, and complete LEXP on three tasks.
    (a) Task success rate over three trials per task and condition.
    (b) Subtask completion rate, calculated as the total number of
    completed subtasks divided by the total number of evaluated
    subtasks across the three trials.
    Labels report successful trials over total trials in (a) and
    completed subtasks over total subtasks in (b).
    Both ablated LEXP variants retain experience-derived feedback;
    the variant without retained experience generates feedback from
    current observations and recent interactions without access to
    stored lessons.}
    \label{fig:ablation}
\end{figure*}

%% file: figures/plan_experience.tex
\begin{figure*}[t]
    \centering
    \begin{minipage}[t]{0.49\textwidth}
        \vspace{0pt}
        \centering
        \includegraphics[width=\linewidth]{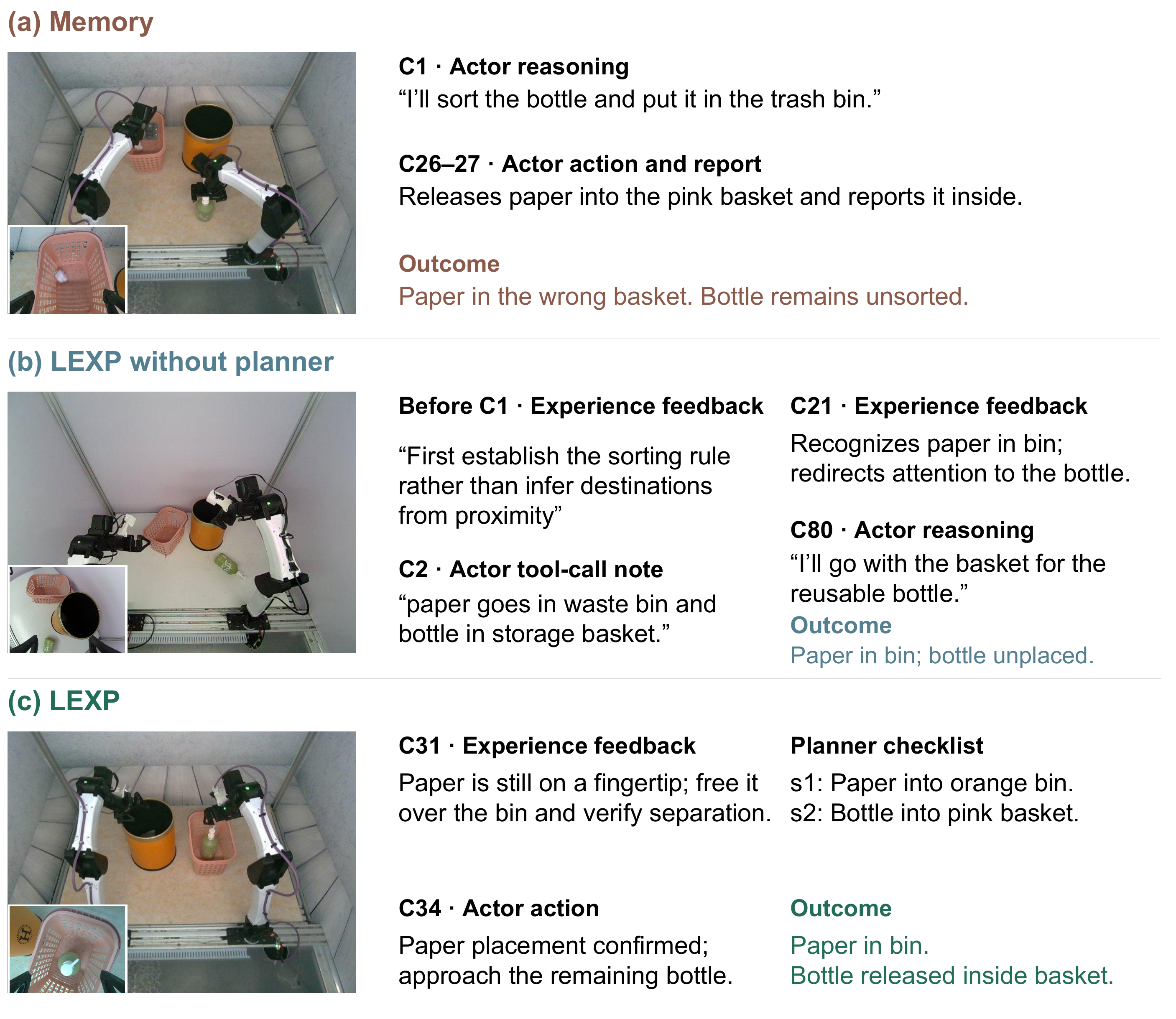}
        \caption{\textbf{Task-level planning in item sorting.}
        Representative reasoning, guidance, and physical outcomes
        with memory, without the planner, and with complete LEXP.}
        \label{fig:planning_sort_comparison}
    \end{minipage}
    \hfill
    \begin{minipage}[t]{0.49\textwidth}
        \vspace{0pt}
        \centering
        \includegraphics[width=\linewidth]{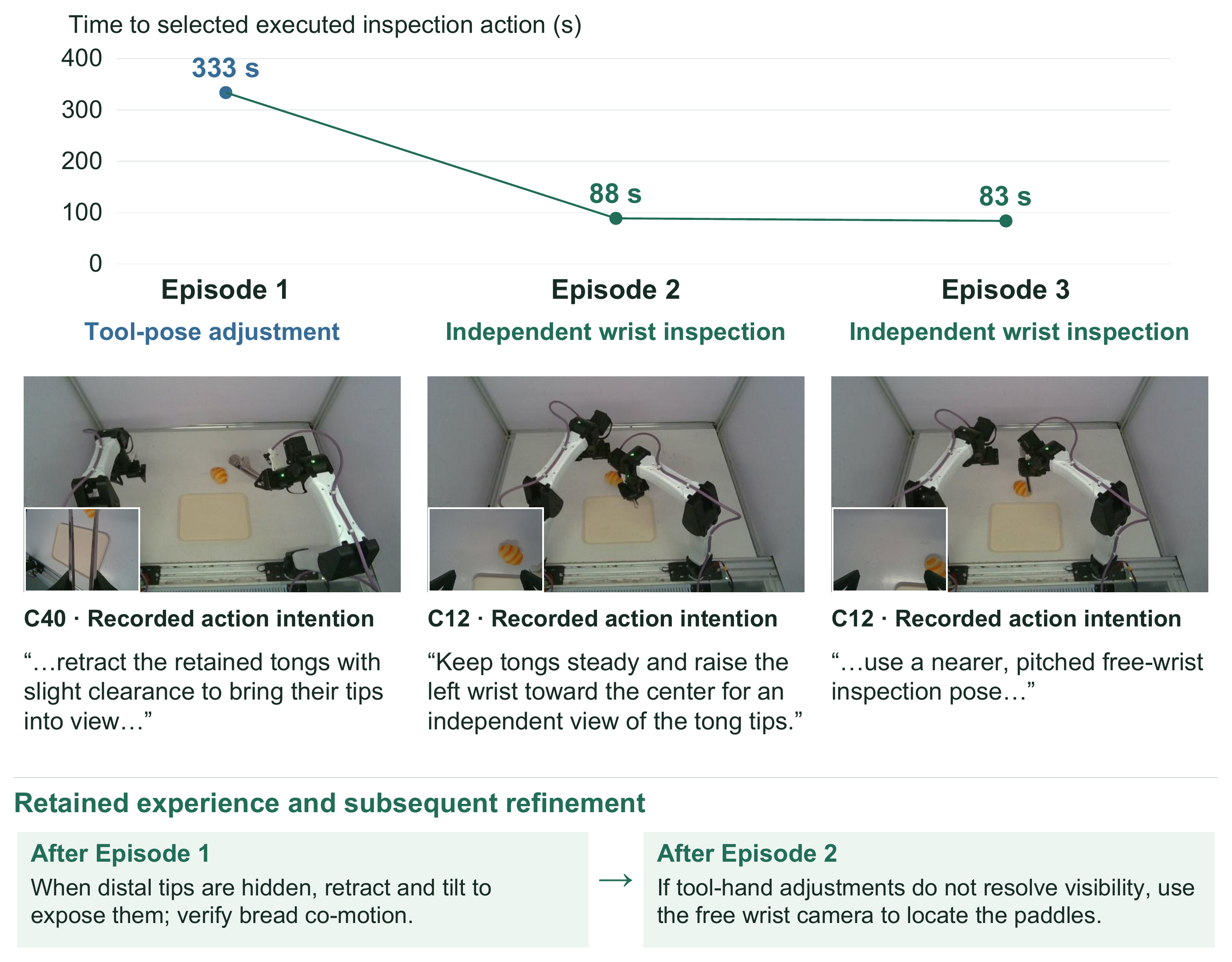}
       \caption{
\textbf{Physical experience becomes reusable guidance in context space.}
Top: elapsed time from the first actor call to the selected executed
inspection action decreases from 333\,s in Episode~1 to 88\,s and
83\,s in Episodes~2 and~3.
}
        \label{fig:experience_reuse}
    \end{minipage}
\end{figure*}

%% file: sections/related_works.tex
\section{Related work}\label{sec:related}

\subsection{Foundation Models for Robotic Manipulation}

Foundation models support robotic manipulation through different interfaces between perception, reasoning, and action. Robot-specialized models such as OpenVLA \citep{openvla} and $\pi_{0.5}$ \citep{pi05} learn action generation from robot data, while DreamZero \citep{dreamzero} and Cosmos Policy \citep{cosmospolicy} connect visual prediction with action generation. These approaches encode manipulation capabilities through model training. A complementary direction uses general-purpose language and multimodal models to reason over tasks and invoke an external action interface. Here, the central challenge is to connect semantic reasoning with physically feasible execution. Different interfaces address this challenge at different levels. SayCan \citep{saycan} grounds language-based task planning in the affordances of available skills, while Code as Policies \citep{codeaspolicies} expresses reasoning through executable programs that call perception and control APIs. VoxPoser \citep{voxposer} translates language and visual observations into spatial value maps for motion generation, and ReKep \citep{rekep} represents manipulation objectives through relational keypoint constraints. Together, these methods show how pretrained reasoning can direct physical behavior through skills, programs, and geometric representations. They also highlight the importance of the surrounding execution system: a model's reasoning must be translated into actions whose consequences can be observed and checked. Recent work increasingly examines this interaction between model capability and execution scaffolding. CaPX \citep{capx} studies code-based manipulation with structured execution feedback and skill synthesis, while concurrent evaluations investigate general-purpose multimodal reasoners as robot policies \citep{gptpolicy,robodojo,robocurve,gptaspolicy}. Our preliminary evaluation follows this direction by examining fine manipulation, spatial reasoning, and responsiveness to context across real-world tasks. Building on these capabilities, LEXP studies how evidence from physical execution can guide subsequent decisions while keeping both model parameters and the action interface fixed.

\subsection{In-context Robot Learning}

In-context robot learning allows behavior to depend on examples supplied at deployment. ICRT \citep{icrt} conditions action prediction on sensorimotor trajectories, enabling demonstrations to specify new behaviors without task-specific parameter updates. RICL \citep{ricl} develops in-context learning capabilities in a pretrained robot policy and retrieves relevant demonstration segments at deployment. RoboPrompt \citep{roboprompt} instead represents demonstrations through textual descriptions of object poses and end-effector actions for an off-the-shelf language model. These approaches demonstrate that contextual examples can communicate task-relevant behavior, although they differ in the representation of demonstrations and the training required to make those examples usable. Feedback extends context conditioning beyond examples supplied before execution. Inner Monologue \citep{innermonologue} incorporates scene descriptions, success feedback, and human input into closed-loop planning, while InCoRo \citep{incoro} investigates in-context control with interaction feedback. Language corrections can also communicate changes that are difficult to specify through a scalar success signal: interactive corrections \citep{sharma2022corrections} and Language Model Predictive Control \citep{liang2024lmpc} use such information to revise robot behavior. These methods establish context as an interface for changing plans and actions during execution. Our preliminary demonstration and language-feedback experiments examine this interface in the evaluated manipulation setting. A further question is how corrective information is obtained and retained. REFLECT \citep{liu2023reflect} uses multisensory execution summaries to explain failures and support corrective planning. DROC \citep{droc} distills human corrections into reusable knowledge and retrieves relevant corrections for subsequent tasks and object instances. These approaches connect immediate feedback with more persistent behavioral guidance. LEXP develops this connection using the robot's own physical interactions: raw history records what happened, while an experience module extracts lessons, retains their applicability conditions and limitations, and interprets them in relation to the current subtask. The resulting feedback brings relevant experience into the actor's context.

\subsection{Learning from Experience and Context Adaptation}

Physical interaction can improve robot behavior through updates to policies, skills, or external knowledge. Reinforcement-learning systems such as SERL \citep{serl} and HIL-SERL \citep{hilserl} improve policies using real-world interaction, with human intervention supporting data collection and recovery. $\pi_{0.6}$ \citep{pi06} likewise studies improvement from experience for generalist robot policies. These approaches update model parameters to incorporate interaction evidence. LEXP pursues improvement through a different adaptation variable: the contextual guidance supplied to a frozen actor. Experience can also modify the structures surrounding a policy. ReSYNC \citep{resync} connects failure recovery with the acquisition of relational concepts for abstract planning. ASPIRE \citep{aspire} refines reusable robot skills through execution evidence, while ENPIRE \citep{enpire} organizes physical policy improvement through an agentic experimentation loop. More recent systems extend adaptation to the execution harness itself. SHAPER \citep{shaper} evolves reusable skills and a context-code harness with frozen model parameters, and Harness VLA \citep{harnessvla} uses execution memory to guide the deployment of fixed manipulation primitives. These works show that experience-driven improvement can operate at several levels beyond policy weights, including programs, abstractions, skills, and contextual knowledge. Related developments in language agents provide mechanisms for retaining and applying experience in textual form. Reflexion \citep{reflexion} uses verbal reflections as memory for subsequent attempts, and ExpeL \citep{expel} extracts reusable insights from accumulated experiences. GEPA \citep{gepa} uses execution traces and reflection to optimize prompts, while Training-Free GRPO \citep{tfgrpo} distills experiential knowledge into contextual guidance. ACE \citep{ace} further organizes adaptation through the incremental generation, reflection, and curation of contextual knowledge. These methods motivate treating context as a persistent adaptation mechanism rather than only a record of the current interaction. Applying this principle to manipulation requires grounding retained guidance in physical evidence. An intended action may not execute as proposed, an executed motion may not achieve its intended effect, and a previously useful correction may depend on object pose, contact state, or task progress. LEXP therefore connects task-level planning with evidence-based experience extraction and reuse. The planner supplies the current task structure; the experience module interprets recent interactions and retained lessons against the observed scene, then generates feedback for the actor. Our evaluation examines how this organization affects real-world manipulation performance, task progression, and the reuse of interaction-derived guidance.

%% file: sections/conclusion.tex
\section{Conclusion}

We propose LEXP, a harness for learning from physical experience
through context-space adaptation. LEXP combines task-level planning
with the extraction and reuse of interaction-derived lessons,
translating physical evidence into contextual feedback without
updating model weights. Across 12 real-world manipulation tasks,
LEXP achieves $23/36$ successful trials (63.9\%), compared with
$19/36$ (52.8\%) with memory alone and $2/36$ (5.6\%) for vanilla
Astra. On three tasks, removing planning reduces success from
$7/9$ to $4/9$, while removing retained experience reduces it to
$3/9$. Execution analyses illustrate how feedback guides action
correction and how retained experience informs a later episode. These results highlight the potential of physical experience as reusable guidance for robotic manipulation.

\newpage

%% file: sections/appendix.tex
\section{Task Details}
\label{app:task_details}

Table~\ref{tab:task_definitions} specifies the instructions, initial configurations, and physical success criteria for the 12 evaluated tasks. Trials across comparison conditions use similar distributions of initial configurations. Task success is determined by human annotation of the observed physical result according to these criteria. Where intermediate progress is reported, subgoals are defined independently of the planner's generated checklist.

\input{tables/task_definition}
\FloatBarrier

\section{Preliminary Context-Steering Setup}
\label{app:context_steering}

The preliminary experiments examine whether externally supplied context can redirect execution and recovery. These inputs are distinct from the agent-generated experience feedback evaluated in LEXP.

\subsection{Visual Demonstrations}

A visual demonstration is a human-recorded video showing how a task is performed. It provides an execution reference, including where present, recovery from unsuccessful interactions. We process each recording into candidate frames, from which the model selects frames relevant to the task. The selected frames are inserted into the actor's context as visual references, alongside observations of the robot's current scene. They condition the actor's decisions without directly specifying robot motion commands.

Specifically, frame selection is performed through a separate, stateless model request containing the task instruction and timestamped candidate images. The model selects a fixed number of frames, including the first and last frames, and provides a relevance explanation for each selection. The frame-selection prompt and the actor's demonstration-input format are shown below.

\begin{promptbox}{Demonstration frame-selection system prompt}
You are preparing to perform a robot manipulation task. Select demonstration frames most useful for completing the task with the robot: task stages, object relationships and before/after effects. Return frame IDs and a short explanation of each frame's relevance. No robot movement is allowed in this phase.
\end{promptbox}

\begin{promptbox}{Frame-selection input template}
TASK: \textless{}task instruction\textgreater{}\par
DEMONSTRATION TYPE: \textless{}demonstration type\textgreater{}\par
Select exactly \textless{}number of selected frames\textgreater{} frames, including endpoints \textless{}first frame ID\textgreater{} and \textless{}last frame ID\textgreater{}.\par
frame \textless{}candidate frame ID\textgreater{}, t=\textless{}timestamp\textgreater{}s\par
Camera: video\par
\textless{}candidate image; repeated for each candidate\textgreater{}
\end{promptbox}

After selected frames,  for human demonstrations, the actor receives the following context block. 

\begin{promptbox}{Visual demonstration in the actor context}
\textless{}HUMAN or RECOVERY\textgreater{} VIDEO DEMONSTRATION: use these selected frames as task references. The demonstrated embodiment and viewpoint may differ. No robot states are supplied in this block. No actions either.\par
human demo frame \textless{}frame ID\textgreater{}, t=\textless{}timestamp\textgreater{}s\par
\textless{}selected image; repeated for each selected frame\textgreater{}
\end{promptbox}

In the towel-folding comparison, clean and recovery recordings provide different execution references; the resulting behavioral differences are reported in Section~\ref{sec:preliminary}.

\subsection{Online Language Feedback}

Online language feedback consists of human-provided textual guidance during robot execution. The feedback is inserted into the actor's context before a subsequent decision, allowing the actor to interpret it together with current observations and generate the corresponding tool call.

Pending messages are inserted into the actor's context using the following wrapper:

\begin{promptbox}{Online language feedback in the actor context}
OPERATOR FEEDBACK:\par
\textless{}human message\textgreater{}\par
\textless{}additional pending messages, if any\textgreater{}
\end{promptbox}

The actor interprets the message together with its current observations and produces its own tool call. 

The barbecue-transfer analysis in Section~\ref{sec:preliminary} relates delivered messages to subsequent changes in motion direction, clearance, arm selection, and action order.

\section{LEXP Implementation Details}
\label{app:implementation}

\subsection{Experience Representation and Updates}

An experience entry contains:
\begin{itemize}
    \item \texttt{lesson}: what was attempted or observed and what followed;
    \item \texttt{applicability}: when the finding applies and what would invalidate it;
    \item \texttt{feedback\_use}: which later decision the finding can inform;
    \item \texttt{limits\_and\_counterevidence}: what the evidence does not establish;
    \item \texttt{unresolved}: alternative explanations or simultaneous changes that remain unresolved;
    \item \texttt{evidence\_ids}: supporting interaction references;
    \item \texttt{revises}: the prior entry replaced by an update, where applicable.
\end{itemize}

The applicability of a retained lesson is distinct from the condition attached to a proposed method. The former concerns reuse of a past finding; the latter concerns whether a current proposal is appropriate.

Experience updates are generated alongside contextual feedback. Each update either creates a new entry or identifies an existing entry through \texttt{revises}. A revision supersedes the referenced entry in the active experience list while preserving its provenance. The resulting list is supplied to subsequent feedback calls. When cross-episode seeding is enabled, findings from earlier attempts are selected and compressed to initialize the new episode's list.

\section{Complete Experimental Results}
\label{app:complete_results}

Table~\ref{tab:full_manipulation_results} reports task-level
success counts for all evaluated models and conditions.
\input{tables/full_manipulation_results}
\FloatBarrier

\section{Prompts and Structured Outputs}
\label{app:prompts}

We provide the prompts and structured-output specifications used for the evaluated configurations.

\subsection{Actor}

The actor system prompt is assembled from embodiment and coordinate conventions, the action contract, configured movement limits, station geometry when supplied, and the verification, retry, and termination instructions. Task instructions, experience feedback, memory, current state and images, and the remaining decision budget are supplied as content. Planner methods and retained lessons reach the actor through experience feedback.

\begin{promptbox}{System-prompt template}
You control a physical robot through one tool call per decision. Observations are measured inputs; action outcomes and task success are distinct. Joint angles are radians; positions are metres in each arm's base frame. Gripper opening is normalized: 0 closed, 1 open. Camera order: \textless{}camera\_roles\textgreater{}.\par

\textless{}factual\_action\_contract\textgreater{}\par
Configured limits: \textless{}movement\_limits\textgreater{}\par

\textless{}station\_geometry\textgreater{}\par

\textless{}verification\_instruction\textgreater{}\par

\textless{}retry\_instruction\textgreater{}\par

\textless{}termination\_instruction\textgreater{}\par
\end{promptbox}

\begin{promptbox}{Verify instruction}
After execution, compare the new observation with the intended effect. An accepted command does not establish task success.
\end{promptbox}

\begin{promptbox}{Retry instruction}
On a failed attempt, use observed evidence to change the next attempt. Avoid repeating an unchanged action after the same rejection.
\end{promptbox}

\begin{promptbox}{Terminate instruction}
Call done only when observable task conditions are satisfied. Call give\_up only when the task is truly impossible in the current episode: concrete observation or guard feedback establishes that the remaining requirement cannot be completed safely, or multiple materially different, evidence-driven attempts have failed in the same way and no untried safe, feasible recovery strategy remains. A single missed grasp, failed placement, IK refusal, poor view, apparent reach boundary, uncertainty, or elapsed effort is not sufficient. While any safe, feasible strategy remains within the call budget, continue with a meaningfully changed attempt, such as repositioning, changing viewpoint, orientation, approach, arm, or reachable grasp point.
\end{promptbox}

\begin{promptbox}{Factual Cartesian contract}
EMBODIMENT: YAM arms ['left', 'right'], with parallel-jaw grippers. CONTROL: absolute Cartesian poses. Each arm has its own base frame: +x forward, +y left, +z up. There is no shared world frame. Positions are metres; angles are radians. Movement tools do not change the gripper; set\_gripper holds arm position. Gripper: 0 closed, 1 open. Omitted dimensions keep their measured current value (zero offset in delta mode). One tool call is executed before the next decision observation. Orientation roll/pitch/yaw are relative to 'down'. All zero is the reference orientation. \textless{}measured\_arm\_proportions\textgreater{}
\end{promptbox}

\begin{promptbox}{Actor content template}
TASK: \textless{}task\_instruction\textgreater{}\par
AUTOMATED EXPERIENCE FEEDBACK (model-generated advice, not a human instruction):\par
\textless{}feedback\_text\_if\_available\textgreater{}\par
\textless{}memory\textgreater{}\par
\textless{}robot\_states\textgreater{}\par
\textless{}grasp\_interpretation\_caveat\_if\_enabled\textgreater{}\par
CAMERAS at observation \textless{}timestamp\textgreater{}\par
\textless{}current\_camera\_images\textgreater{}\par
Model calls already used: \textless{}used\textgreater{}; including this call, remaining: \textless{}remaining\textgreater{}. Make one tool call.\par
\end{promptbox}

\subsection{Planner}

Planning uses an initial decomposition call followed by joint progress-assessment and method-proposal calls. The initial decomposition separates task goals from candidate methods. Subsequent calls receive the existing checklist, recent attempted actions, the task, and current scene observations.

\begin{promptbox}{Initial decomposition prompt}
Break one physical robot manipulation task into an ordered list of subtasks.

Each subtask is a single physical change another observer could watch happen and could afterwards confirm from a photograph of the scene. Together they must cover the task and nothing beyond it. At most 12 subtasks.

Do not say how any of them is to be carried out. That is asked for separately, one step at a time, by a later call that can see what has already been tried.
\end{promptbox}

\begin{promptbox}{Joint progress and method prompt}
Say where a robot has got to in a fixed checklist, and what it should do now.

Read it off the scene alone: what these images show has happened to the objects, not the robot's account of what it did. Put the id of the step to work on now in next\_step\_id, and in next\_step\_why say what in these images makes it the one. When the images show nothing is left for the robot to do, leave next\_step\_id empty and say what shows the task is over. A step whose effect the scene was never going to keep is not work that remains: once what it led to is there, it is behind the robot.

Everything before that step counts as passed, everything from it on as still to do, and you may not reword a step that is already there. If the scene has gone backwards -- something dropped, something undone -- name the step that has to be done again; that is how going backwards is said here.

When the step in play moves forward, say in passed\_evidence what these images show about the one just passed. Leave it empty when the step in play has not changed, and empty when it has gone back: that reading is made once, where it happened, and is not written again afterwards.

The checklist itself may gain or lose a step, and only in those two ways. ADD one when this scene shows something no step accounts for -- say in why what you are looking at, not that a step seems missing. RETIRE one when the scene shows it is not needed after all; it stays on the list, marked, because that it was planned and found unnecessary is worth knowing. You may not retire a step the run has already passed. Both lists empty is the ordinary answer.

Return two separate outputs. First judge the checklist under plan. Then propose current\_method for the step you name in next\_step\_id. The method goes to an advice module, not directly to the robot. A proposed way of acting is not an extra goal or evidence of success; never add a subtask merely to require your method. Keep goals and means distinct. Describe what contacts, supports or moves relative to what, without coordinates, joint angles or gripper commands. State its applicability condition (use 'none' if none) and expected observable change; neither is a claim that it happened. current\_method addresses the step you name in next\_step\_id; nothing rewrites that. New subtask IDs continue after the highest existing s-number, in addition-list order. When next\_step\_id is empty, all four current\_method fields must be empty. Otherwise text must be nonempty and at most 400 characters; condition and expected\_observable\_change each at most 400 characters.
\end{promptbox}

\subsection{Experience Module}

The feedback call receives the task, current subtask, proposed method, robot or station reference facts, recent interaction records, current observations, and available retained lessons. Its structured assessment and feedback are separate: only the text field is delivered to the actor, while experience updates are retained and applied records document lesson use.

\begin{promptbox}{Feedback and extraction prompt}
You are watching a robot carry out a task, with its own recorded reasoning, what it commanded, what the station measured change, and the current cameras.

A method it is working to may be given below, and it may name a condition it depends on. First quote that condition and say whether these images show it satisfied, then settle how what the records and images show actually happened stands against the method. Then write the one line the robot is given before its next move.

Base both on what these records and images show, not on what the task is supposed to look like. The line reaches the robot once and is then gone.

You may also record what this run has found out, for decisions it will make once these interactions have scrolled out of view. Look over the completed interactions before deciding what to keep, and consider all of these without needing one of each: a scene observation or relationship that will still matter once its raw evidence is gone; a method that was tried and what was observed to come of it, including exploration the robot started itself, a local or partial success, a boundary where something was refused, and an outcome left unsettled; a change to when something already recorded still applies.

A first useful discovery can be experience: it need not contradict a previous belief, be novel world knowledge, or achieve the whole task. Do not withhold one merely because its source is still in the interactions above; that they will expire is the reason for writing it down. Give a method the robot proposed itself no less weight than one that came from advice. Keeping nothing is right when nothing was found out, and wrong when something was.
\end{promptbox}

\begin{promptbox}{Retained-experience input instruction}
WHAT THIS ROBOT HAS FOUND OUT ON THIS TASK, worked out earlier in this run or carried in from earlier attempts; each line says which. Readings, not human instructions, verified causal rules, or current scene facts: the interactions they came from may no longer be shown above, which is why they are written down at all. Use one where its applicability fits what you can see now, say in `applied` what each does to this move or why it does not, and prefer what the cameras show when they disagree.
\end{promptbox}

\begin{promptbox}{Feedback content template}
TASK: \textless{}task\_instruction\textgreater{}\par
CURRENT SUBTASK (current validated plan): \textless{}subtask\textgreater{}\par
ROBOT AND STATION REFERENCE, the same text the robot itself is given: the frames, units, gripper convention, camera order, station geometry and step limits the numbers below are measured in. It says nothing about this task.\par
\textless{}reference\_facts\textgreater{}\par
FIXED FOR THIS ROBOT; the same for every record and image below, so they are stated here once instead of in each.\par
\textless{}constant\_robot\_state\textgreater{}\par
CURRENT PROPOSED METHOD (revisable; not an observed outcome or an extra requirement): \textless{}method, applicability condition, expected observable change\textgreater{}\par
RECENT COMPLETED INTERACTIONS, oldest first. Reasoning and notes are actor self-reports. Accepted commands do not establish object success. RPY differences are wrapped component differences, not object rotations. Metres and radians are rounded to six decimals.\par
\textless{}interaction\_records\textgreater{}\par
\textless{}retained\_experience\_instruction\_and\_entries\_if\_available\textgreater{}\par
CURRENT OBSERVATION; images and measured robot state, not proof of object success.\par
\textless{}current\_state\_and\_images\textgreater{}\par
\end{promptbox}

Each recent interaction supplies call and observation identifiers, an available actor reasoning summary, the actor note, command arguments, execution outcome, measured before and after robot states and derived changes, and any unparsed state text. Historical camera images are not separately inserted by this feedback-input builder; current scene images are supplied. Optional blocks are absent when their inputs are unavailable.

\paragraph{Cross-episode initialization.} When experience seeding is enabled, a separate call selects and compresses findings from earlier attempts into the lesson list used to initialize the new episode. This call reads retained findings, rather than the complete source trajectories.

\begin{promptbox}{Experience-seeding prompt}
Below is what earlier attempts at this task on this station found out: for each, what was observed, when it applies, which decision it can constrain, what it does not establish, and anything it could not tell apart.
Write what this episode should start out already knowing. It faces a different arrangement of the same task on the same station, so carry only what will still be true there -- an applicability that names this one arrangement cannot be used in another. Keep what each entry does not establish; dropping it is how a local result becomes a rule.
Everything you carry costs this episode something to read, and carrying nothing is a legitimate answer.
\end{promptbox}

%% file: tables/task_definition.tex
\begingroup
\small
\setlength{\tabcolsep}{5pt}
\renewcommand{\arraystretch}{1.15}
\setlength{\LTleft}{0pt}
\setlength{\LTright}{0pt}
\setlength{\LTcapwidth}{\linewidth}
\begin{longtable}{@{}
  >{\raggedright\arraybackslash}p{0.17\linewidth}
  >{\raggedright\arraybackslash}p{0.20\linewidth}
  >{\raggedright\arraybackslash}p{0.25\linewidth}
  >{\raggedright\arraybackslash}p{\dimexpr0.38\linewidth-6\tabcolsep\relax}
@{}}
\caption{Task instructions, initial configurations, and physical success criteria for the 12 real-world manipulation tasks.}
\label{tab:task_definitions}\\
\toprule
\textbf{Task} & \textbf{Instruction} & \textbf{Initial configuration} & \textbf{Success criterion} \\
\midrule
\endfirsthead
\multicolumn{4}{@{}l}{\small\tablename~\thetable\ (continued)}\\
\toprule
\textbf{Task} & \textbf{Instruction} & \textbf{Initial configuration} & \textbf{Success criterion} \\
\midrule
\endhead
\midrule
\multicolumn{4}{r@{}}{\small Continued on next page}\\
\endfoot
\bottomrule
\endlastfoot

Pick up a ping-pong ball
& Pick up the ball.
& A ping-pong ball rests inside a shallow transparent dish on the tabletop; both arms start parked clear of the dish.
& The ball is enclosed between the fingertips and held clear of the dish, which remains on the tabletop. \\

Pick up a cable tie
& Pick up the cable tie.
& A white cable tie lies on an orange tray on the tabletop.
& The cable tie is retained between the fingertips, with its free length extending from the raised gripper clear of the tray. \\

Pick up a key
& Pick up the key.
& A key on a split ring with an attached tag lies on the table.
& The key, ring, and tag are suspended from the raised gripper clear of the surface; the grasp may be on the key, ring, or tag. \\

Pick up an L-shaped hex key
& Pick up the hex key.
& An L-shaped hex key lies on the table with its long leg and bent end accessible.
& The hex-key shaft is pinched between the opposing gripper pads, with its bent end projecting clear of the work surface. \\

Stand up a bottle
& Stand the bottle upright.
& The bottle lies on its side on the tabletop.
& The bottle stands cap-up on its base on the tabletop, with visible clearance from both grippers and no robot support. \\

Grasp out of reach plush
& Pick up the plush toy.
& A plush toy rests on a mat positioned beyond the arm's direct reach.
& Plush fabric is retained between the fingertips, and the toy hangs from the raised gripper clear of the mat, which remains flat below. \\

Transfer bread with tongs
& Use the tong to pick up all the bread and put on the tray.
& Tongs rest on the tabletop; bread pieces lie on the table behind a tray.
& All bread pieces are visible within the tray in the overhead view, while the tongs' working ends are separated, empty, and clear of the bread. \\

Move a potato chip intact
& Pick up the chip.
& A potato chip rests on the tabletop.
& The chip is pinched between the fingertips, suspended clear of the surface, and remains unbroken. \\

Sweep up trash
& Sweep the debris into the dustpan.
& Debris lies on the work surface; a brush and a dustpan rest nearby, to be held one per arm.
& One gripper holds the brush handle and the other holds the dustpan. The dustpan's lip rests against the surface with its mouth facing the debris, and the debris ends up inside the pan. \\

Fold a towel in half
& Fold the towel in half.
& A towel lies flat and unfolded on the table.
& The towel rests as a doubled rectangle with one half covering the other, with opposite corners and side hems brought together. \\

Sort trash and toiletries
& Sort the items on the table into the appropriate containers.
& Crumpled paper and a pump bottle lie on the table beside a cylindrical bin and a basket.
& The crumpled paper rests inside the bin, and the pump bottle stands on the floor of the basket, both separate from the withdrawn fingers. \\

Retrieve drawer
& Find the drawer with the sticker. Open each drawer at most once.
& A three-drawer cabinet with knobs starts with all drawers shut; a sheet of stickers is inside exactly one drawer.
& A drawer is open with the sticker sheet visible inside. Each drawer has been opened at most once, and non-target drawers have been pushed shut again. \\

\end{longtable}
\endgroup

%% file: tables/full_manipulation_results.tex
\begin{table}[!htbp]
\centering
\small
\setlength{\tabcolsep}{4pt}
\renewcommand{\arraystretch}{1.15}
\caption{\textbf{Complete task-success results.} Entries report successful trials over total trials. Overall results use all 12 tasks, with three trials per task and condition.}
\label{tab:full_manipulation_results}
\begin{tabular}{lccccc}
\toprule
& \multicolumn{3}{c}{GPT-6 Astra}
& Opus 5.5
& Sol 5.6 \\
\cmidrule(lr){2-4}\cmidrule(lr){5-5}\cmidrule(lr){6-6}
Task & Vanilla & Memory & LEXP & Memory & Memory \\
\midrule
Pick up a ping-pong ball & 2/3 & 2/3 & \textbf{3/3} & 1/3 & 0/3 \\
Pick up a cable tie & 0/3 & \textbf{3/3} & 2/3 & 1/3 & 0/3 \\
Pick up a key & 0/3 & \textbf{3/3} & \textbf{3/3} & 1/3 & 0/3 \\
Pick up an L-shaped hex key & 0/3 & \textbf{2/3} & \textbf{2/3} & \textbf{2/3} & 0/3 \\
Stand up a bottle & 0/3 & \textbf{2/3} & \textbf{2/3} & 0/3 & 0/3 \\
Grasp out of reach plush & 0/3 & \textbf{3/3} & \textbf{3/3} & \textbf{3/3} & 0/3 \\
Transfer bread with tongs & 0/3 & 2/3 & \textbf{3/3} & 1/3 & 0/3 \\
Move a potato chip intact & 0/3 & \textbf{1/3} & \textbf{1/3} & 0/3 & 0/3 \\
Sweep up trash & 0/3 & 0/3 & 0/3 & 0/3 & 0/3 \\
Fold a towel in half & 0/3 & 0/3 & 0/3 & 0/3 & 0/3 \\
Sort trash and toiletries & 0/3 & 1/3 & \textbf{2/3} & 0/3 & 0/3 \\
Retrieve drawer & 0/3 & 0/3 & \textbf{2/3} & \textbf{2/3} & 0/3 \\
\midrule
Overall (12 tasks) & 2/36 & 19/36 & \textbf{23/36} & 11/36 & 0/36 \\
Success rate (\%) & 5.6 & 52.8 & \textbf{63.9} & 30.6 & 0.0 \\
\bottomrule
\end{tabular}
\end{table}